\documentclass[letterpaper, 10 pt, conference]{ieeeconf}  

\IEEEoverridecommandlockouts                              

\usepackage{graphics}
\usepackage{amsmath}
\usepackage{graphicx}
\usepackage{floatrow}
\usepackage{layout}
\usepackage{amssymb} 
\usepackage{multirow}
\usepackage{caption}
\usepackage{indentfirst}
\usepackage{caption}
\usepackage{subcaption}
\usepackage{bbold}

\newcommand{\RN}[1]{%
  \textup{\uppercase\expandafter{\romannumeral#1}}%
}

\title{\LARGE \bf
Explaining the Decisions of Deep Policy Networks for Robotic Manipulations
}

\author{Seongun Kim$^{1}$ and Jaesik Choi$^{1}$
\thanks{$^{1}$ Graduate School of Artificial Intelligence, Korea Advanced Institute of Science and Technology, 291 Daehak-ro, Daejeon 34141, Republic of Korea. Correspondence to: Jaesik Choi $<$jaesik.choi@kaist.ac.kr$>$.}
}

\begin{document}

\maketitle
\thispagestyle{empty}
\pagestyle{empty}
\IEEEpeerreviewmaketitle

\begin{abstract}

Deep policy networks enable robots to learn behaviors to solve various real-world complex tasks in an end-to-end fashion. However, they lack transparency to provide the reasons of actions. Thus, such a black-box model often results in low reliability and disruptive actions during the deployment of the robot in practice. To enhance its transparency, it is important to explain robot behaviors by considering the extent to which each input feature contributes to determining a given action. In this paper, we present an explicit analysis of deep policy models through input attribution methods to explain how and to what extent each input feature affects the decisions of the robot policy models. To this end, we present two methods for applying input attribution methods to robot policy networks: (1) we measure the importance factor of each joint torque to reflect the influence of the motor torque on the end-effector movement, and (2) we modify a relevance propagation method to handle negative inputs and outputs in deep policy networks properly. To the best of our knowledge, this is the first report to identify the dynamic changes of input attributions of multi-modal sensor inputs in deep policy networks online for robotic manipulation.

\end{abstract}

\section{INTRODUCTION}
\label{introduction}

There has recently been substantial progress in representing robot policies using deep neural networks (DNNs). DNNs enable a robot policy to take high dimensional sensor data as an input, which makes the policy sensorimotor. Combined with the reinforcement learning method, deep policy networks have shown that they can outperform humans in actions such as playing board games \cite{alpha_go, alpha_zero} or video games \cite{atari, alpha_star}. However, due to the high dimensional action space, learning a close-to-optimal policy which solves complex manipulation tasks with real-world robots is still challenging.

Over the last few years, it has been proven that a deep visuomotor policy can solve complex tasks in real-world robots. These policies have succeeded in assembling a toy airplane \cite{gps_assembling}, placing a coat hanger on a rack \cite{gps_visuomotor}, hitting a puck into a goal with a stick (hockey) \cite{gps_hockey}, and opening a door [5]. In spite of these successes, the internal mechanisms of deep visuomotor policies have not been clearly analyzed due to the lack of transparency. As the demand for reliable intelligent robot systems increases, it is essential to understand how policy networks work and which role each input feature is responsible for. It has been implicitly shown that convolutional layers are responsible for perception and that fully-connected layers are responsible for control \cite{gps_modular}. However, it has not been shown to what extent each feature contributes to the end-effector movement and how each input feature affects the decision making of the robot policy models quantitatively.

Recently, much work has focused on implementing DNN models with explainability. Methods to discover some internal nodes which are highly related to human-interpretable semantic concepts were introduced for convolutional neural network (CNN) models \cite{network_dissection} and generative adversarial network (GAN) models \cite{gan_dissection, e_gbas}. Several input attribution methods \cite{dtd, lrp, lrp_nlp, integrated, grad_cam, rap, deep_lift} provide intuitive explanations in classification problems. They backpropagate outputs to each input features to measuring the attributions of each input feature quantitatively.

\begin{figure*} [tb]
    \centering
    \includegraphics[width=\linewidth]{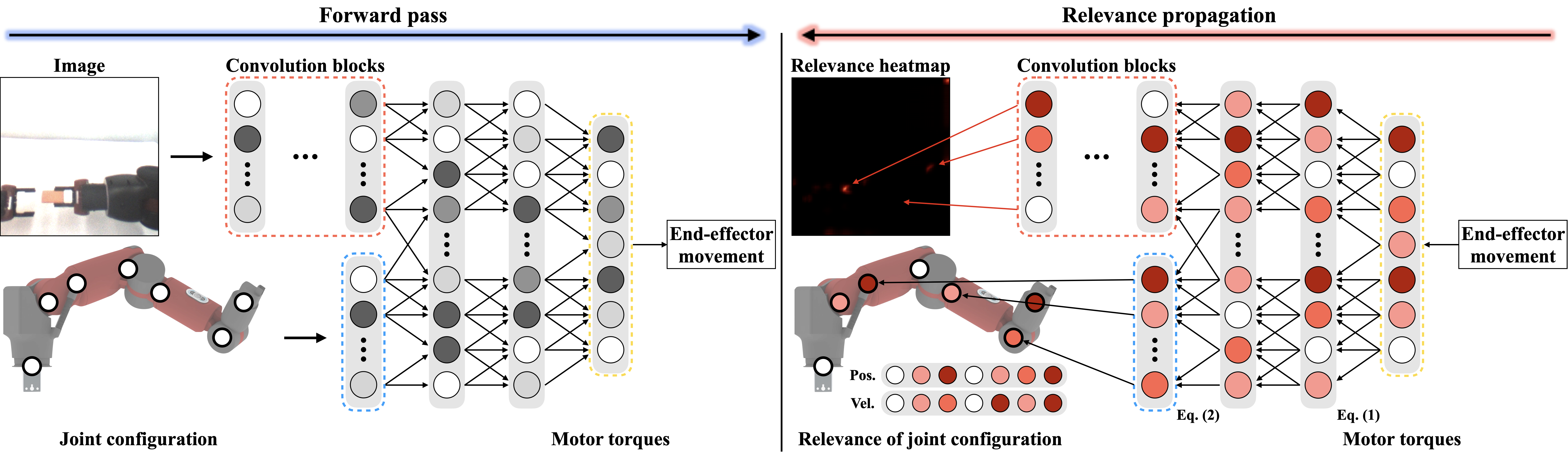}
    \caption{Deep policy network architecture. The forward pass directly outputs the motor torques from the visual input. Feature points which are derived from spatial softmax layer and represent locational information in the image space, are concatenated with the joint configuration for the input of a fully-connected layers. The backward pass redistributes the relevance from each motor torque to the visual input and the configuration input with a layer-wise relevance propagation method. The redistributed relevance is added in proportional to the degree to which each motor torque contributes to the joint or end-effector movement. The heatmap of each input represents the extent to which each input attributes to the motor torques.
}
    \label{fig:architecture}
\end{figure*}

However, input attribution methods proposed for classification networks are not directly applicable to robot policy networks in two reasons. First, an end-effector does not move in direct proportion to each joint torque. The number of links from the joint affects to the end-effector movement. Second, unlike most image classification problems, there exist negative inputs and outputs in robotic manipulation problems. The existence of negative values leads to relevance degradation and breaks some assumptions that should be met when deriving the above input attribution methods. To address these problems, we propose a method that measures the importance factor for each joint torque and modifies relevance propagation rules for the proper handling of negative inputs and outputs.

In this paper, we present several new empirical analyses of deep policy networks for robotic manipulation tasks. We utilize existing input attribution methods with the proposed modification methods for robot policy networks. With these methods, we look inside the model and measure the extent to which each input attributes to the motor torque and the end-effector movement. To verify the correctness of our experimental results, we apply three different input attribution methods: Deep Taylor Decomposition \cite{dtd}, Relative Attributing Propagation \cite{rap} and Guided Backpropagation \cite{gbp}.

We qualitatively analyze deep policy networks by visualizing a heatmap and identifying whether the model actually focuses on the input features as expected and how it affects the performance. We quantitatively analyze the behavior of the policy model and identify how the input features affect the decisions of the model. To the best of our knowledge, this is the first report to capture the dynamic changes of sensors' contributions in deep policy networks for robotic manipulations.

\section{Background}
\label{sec:background}

\subsection{Input Attribution Methods}
An \emph{input attribution method} is a method for measuring the contributions of input features to the decisions of DNNs. There are several ways to measure these contributions, which we call the \textit{relevance}. One way is to decompose the DNN into a set of linear functions by approximating it with linear functions \cite{dtd, lrp, lrp_nlp}. Another line of recent research uses heuristic rules that generate a clear distinction between the target object and the background \cite{rap} or addresses two problems: the saturation problem and the thresholding problem \cite{deep_lift}. It has been also shown that the gradient of the output with respect to the input can be viewed as the attributions \cite{integrated, grad_cam, gbp, grad_input}. All methods listed above propagate the relevance in a backward manner from the outputs to the inputs. However, in spite of a solid theoretical foundation of decomposing methods \cite{dtd}, there is no ground truth value for the relevance \cite{dtd_theoretical_1, dtd_theoretical2}. Therefore, to validate the reliability of the results of this paper, we utilize three different methods: one that decomposes the DNN, one that uses the heuristic propagation rules, and one that uses the gradients. Detailed explanations of each are given below.

\subsection{Deep Taylor Decomposition}
Deep Taylor Decomposition (DTD) \cite{dtd} is an input attribution method which utilizes Taylor decomposition layer by layer in a backward manner in DNNs. The complex nonlinear model $f(x)$ represented by the DNN is a function which maps high dimensional inputs to outputs. It can be approximated with the first order Taylor expansion given a data point $\tilde{x}$. If $\tilde{x}$ is some well-chosen root point which makes the function value 0, the approximated nonlinear function $f(x)$ can be viewed as a linear function centered at zero. If $\tilde{x}$ lies in a high dimensional space, the approximated function can be represented as a set of linear functions, with each linear function mapping each dimension's input to the output. Therefore, each approximated function value is considered as the contribution of the corresponding input to the output.

However, for complex nonlinear models such as DNNs, it is problematic to approximate them with the first order Taylor expansion since it is difficult to find the root point and higher-order terms are ignored. Therefore, DTD approximates a layer-to-layer mapping with first order Taylor expansion, which enables one to decompose layer-wise mapping and compute the contributions of the layer-wise inputs which come from the outputs of previous layers.

Depending on the choice of the root points of layer-to-layer mappings, multiple propagation rules can be derived. In this paper, we use $z^+$-rule which is one of the representative propagation rules. The $z^+$-rule is obtained by choosing the nearest root on the segment $(\{x_i\mathbb{1}_{w_{ij}<0}\}, \{x_i\})$, where $\{x_i\}$ is the given data point and $w_{ij}$ is the weight. $z_{ij}^+$ is defined as $x_i w_{ij}^+$, where $w_{ij}^+$ is defined to be $\max(w_{ij}, 0)$.

\subsection{Relative Attributing Propagation}
Relative Attributing Propagation (RAP) \cite{rap} is an input attribution method with rule-based relevance propagation. RAP provides a way of handling negative attributions by separating attributions into two types: relevant and irrelevant, according to the relative influence between the layers. Instead of ignoring the negative propagated relevance, it measures the absolute influence by taking absolute to all propagated relevance. It then normalizes these absolute relevance scores and propagates them again to the previous layer. By considering highly relevant and highly irrelevant influence at the same time, it provides a much clearer and more intuitive heatmap than other methods in image classification problems.

\subsection{Guided Backpropagation}
Guided Backpropagation (GBP) \cite{gbp} is a pure gradient-based attribution method. It produces a clear visualization result (i.e., a heatmap) by utilizing a guidance signal with its gradient. Computing the gradient of the output with respect to the input produces a heatmap conditioned on the input image. However, simply computing the gradient does not provide clear visualization, as the negative gradient can be backpropagated.

This problem is resolved by utilizing a guidance signal when following a backward pass. The guidance signal is achieved by combining backpropagation and deconvnet \cite{deconvnet}. The gradient signal is zeroed out for negative activations by backpropagation when ReLU-like activation functions are applied. Similarly, the gradient is zeroed out for the negative gradient by deconvnet. Combining these two, Guided Backpropagation retains only positive gradient signals: $R_i = \sum_j \mathbb{1}_{x_i>0} \cdot \mathbb{1}_{\frac{\partial R_j}{\partial x_i} > 0} \cdot \frac{\partial R_j}{\partial x_i}$, where $R_i$ and $R_j$ are the relevance of the $i$-th node at layer $\{i\}$ and the $j$-th node at layer $\{j\}$, respectively, and $\mathbb{1}_{\{\cdot\}}$ is the indicator function.
\section{Explaining Deep Policy Networks with Multiple Sensors in Robotic Manipulation Tasks}
\label{sec:explaining_models}
We explain how the deep policy networks make the decision in the form of the motor torque by analyzing the contributions of the input sensors. Specifically, the deep policy networks represented by the neural network shown in Fig. \ref{fig:architecture} is employed, which is trained with the Guided Policy Search (GPS) algorithm \cite{gps_visuomotor}. First, we assign an importance factor $\alpha_j$ to a motor torque $\tau_j$ in proportion to the degree to which each motor torque contributes to the end-effector movement. The importance factor $\alpha_j$ measures how far the end-effector moves by applying the joint torque $\tau_j$ to each joint $j$.

\begin{figure} [tb]
  \includegraphics[width=\linewidth]{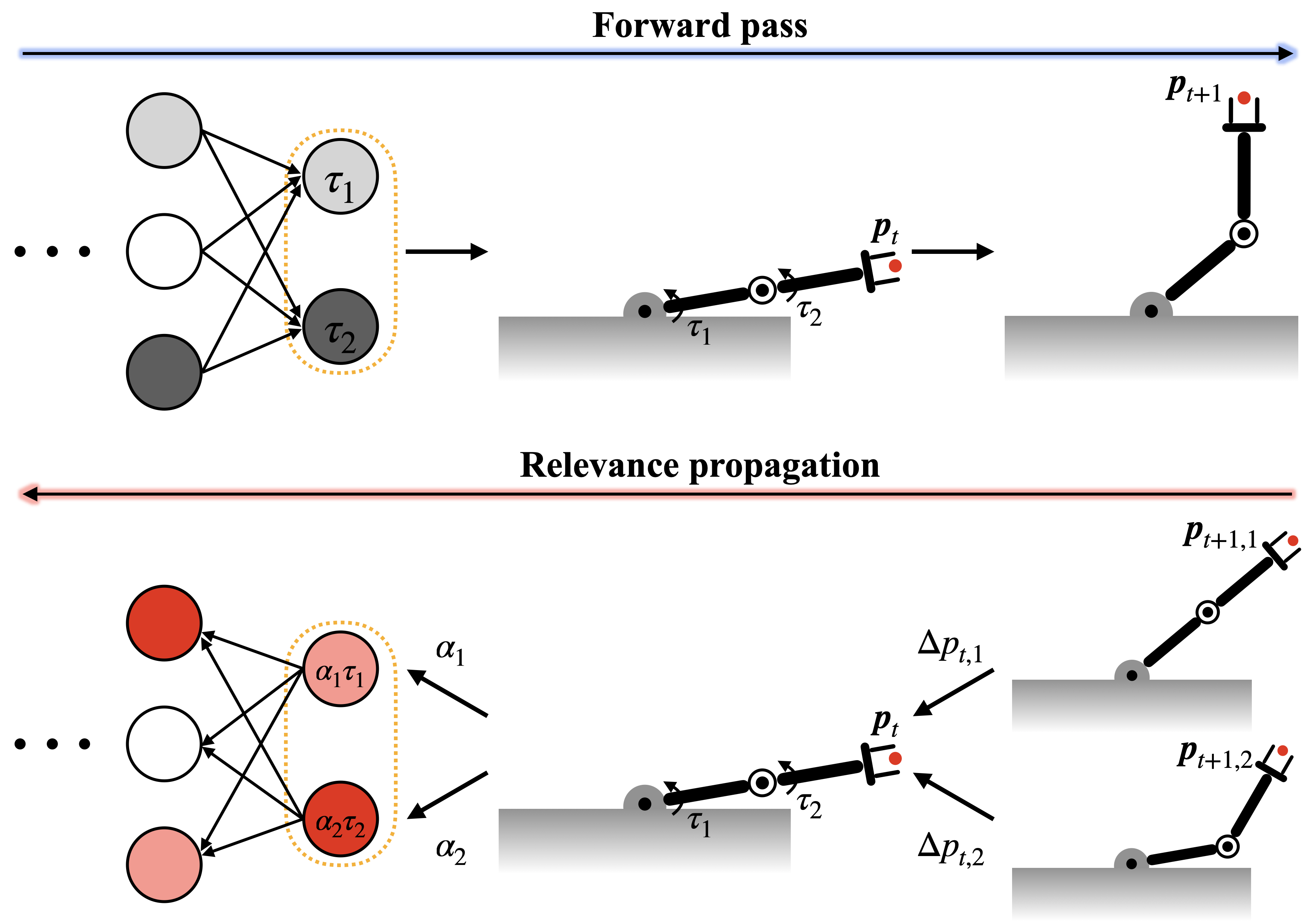}
    \caption{An illustration of relevance propagation at the output layer. The motor torque outputs are weighted in proportion to the degree to which each motor torque contributes to the end-effector movement.}
    \label{fig:multiple_outputs}
\end{figure}

We compute the importance factor $\alpha_j$ of each joint $j$ by measuring the end-effector movement when applying each joint torque $\tau_j$ separately using the dynamics and forward kinematics, as shown in Fig. \ref{fig:multiple_outputs}. In Fig. \ref{fig:multiple_outputs}, $\alpha_j$ is defined to be proportional to $\Delta \boldsymbol{p}_{t, j} / \tau_j$, where $\Delta \boldsymbol{p}_{t, j}$ is the end-effector movement at time step $t$ when only applying joint torque $\tau_j$. If the dynamics and forward kinematics are unavailable, $\alpha_j$ is set to $1$ under the assumption that the same amount of torque moves the same angle of the joint. The amortized input attribution is the $\alpha_j$ weighted sum of the input attributions computed from the individual outputs of each joint. 

To explain this further, we utilize three different input attribution methods (DTD, RAP and GBP) to validate the reliability of the result. However, DTD and RAP implicitly compute the gradient $\times$ data point to include the global effect in the explanation, while GBP only computes the gradient explicitly, only measuring a local effect of the input features on the decision of the models. Therefore, we multiply the input data point by the relevance achieved from GBP for a fair comparison of the three different methods. 

Directly applying attribution methods to robot policy models is not plausible due to negative values in the inputs and outputs. Existing input attribution methods assume that the inputs and outputs of DNNs are positive. As an example, in image recognition tasks, it is possible to make all input pixel values and output (classification) values positive. However, in the robotic manipulation tasks, it is difficult to make all inputs and outputs positive because robot configurations and motor torques are zero-centered values.

The coexistence of positive outputs and negative outputs causes the relevance degradation problem. As the ReLU activation function is applied to all layers except for the last layer, 
if the motor torque is negative, the relevance is also negative as the gradient is negative while the input is positive. When adding all positive relevance values and negative relevance values in the last hidden layer, the relevance degradation problem occurs. Moreover, the root point is not guaranteed to be found when applying the $z^+$-rule to the negative output. 
\begin{figure*}
    \centering
    \begin{subfigure} {.35\textwidth}
    \centering
    \includegraphics[width=\linewidth]{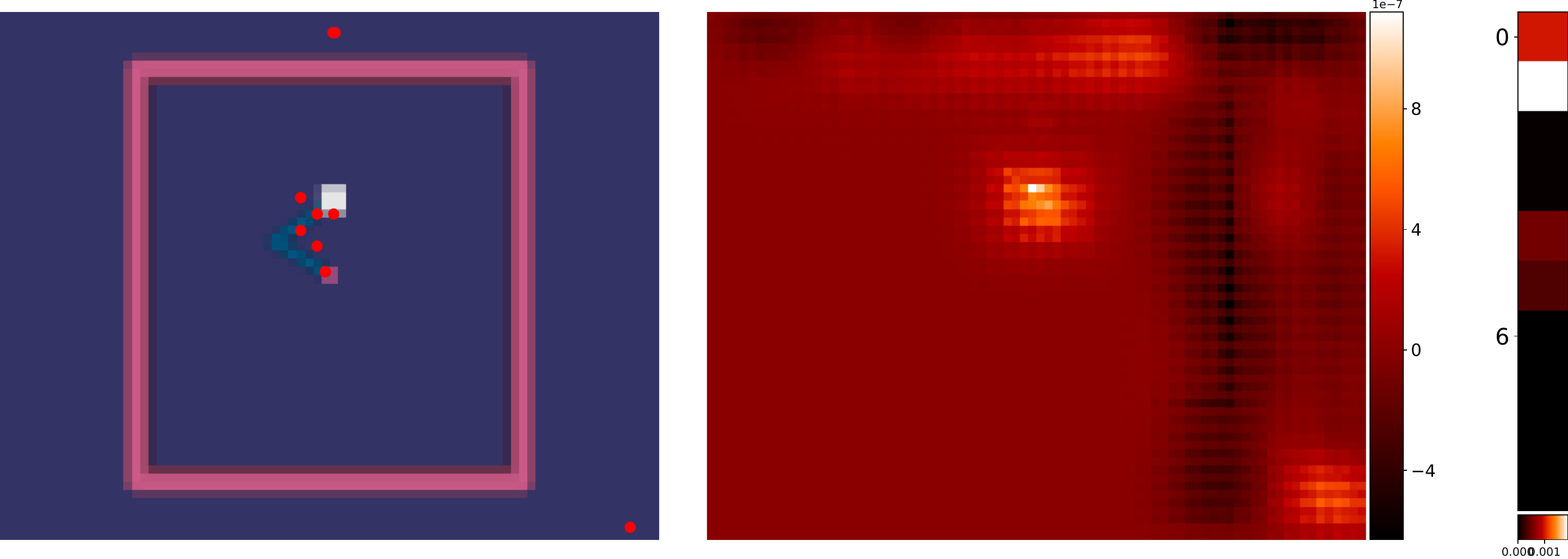}
    \caption{A heatmap in the MuJoCo reaching task}
    \label{fig:heatmap_mjc_reacher}
    \end{subfigure} \hspace{0.1\textwidth}
    \begin{subfigure} {.35\textwidth}
    \centering
    \includegraphics[width=\linewidth]{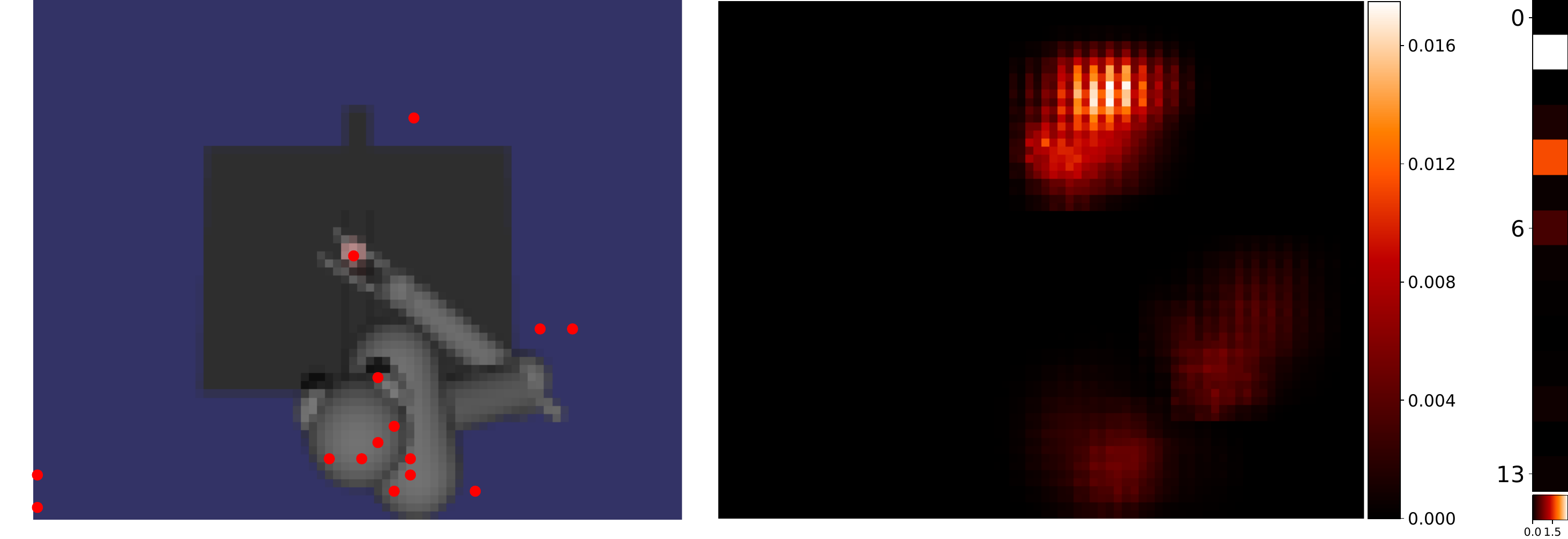}
    \caption{A heatmap in the MuJoCo peg insertion task}
    \label{fig:heatmap_mjc_peg}
    \end{subfigure}
    \begin{subfigure} {.35\textwidth}
    \centering
    \includegraphics[width=\linewidth]{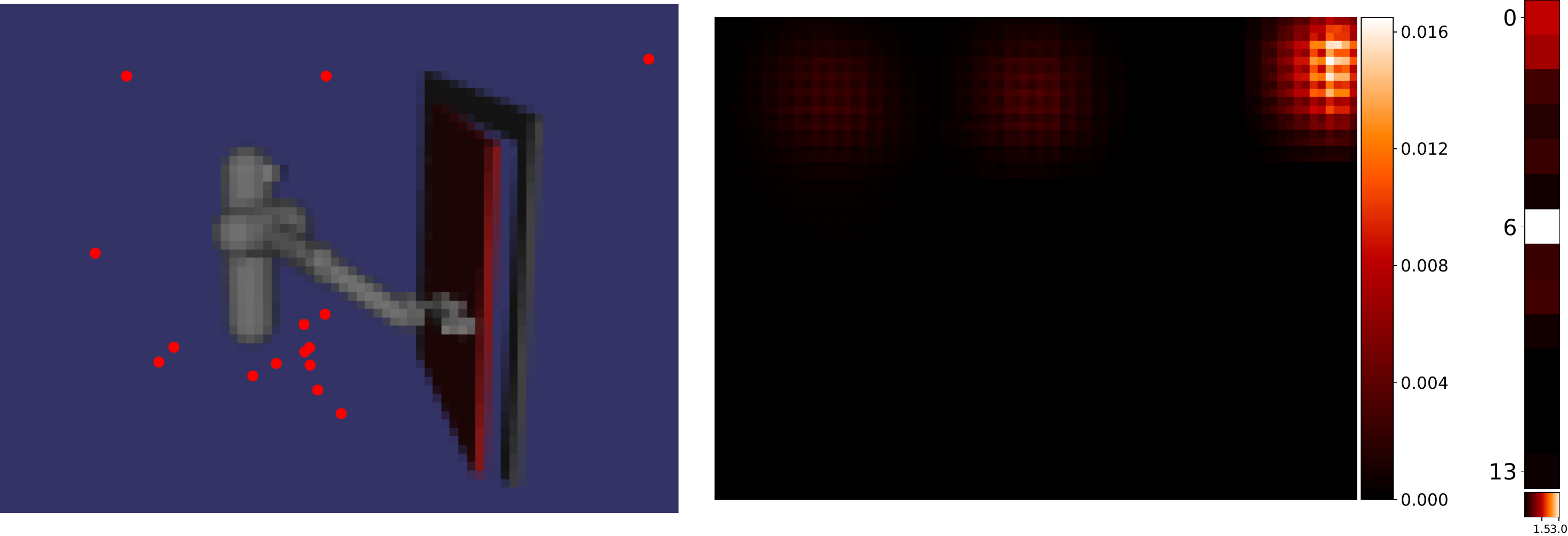}
    \caption{A heatmap in the MuJoCo door opening task}
    \label{fig:heatmap_mjc_door}
    \end{subfigure} \hspace{0.1\textwidth}
    \begin{subfigure} {.35\textwidth}
    \centering
    \includegraphics[width=\linewidth]{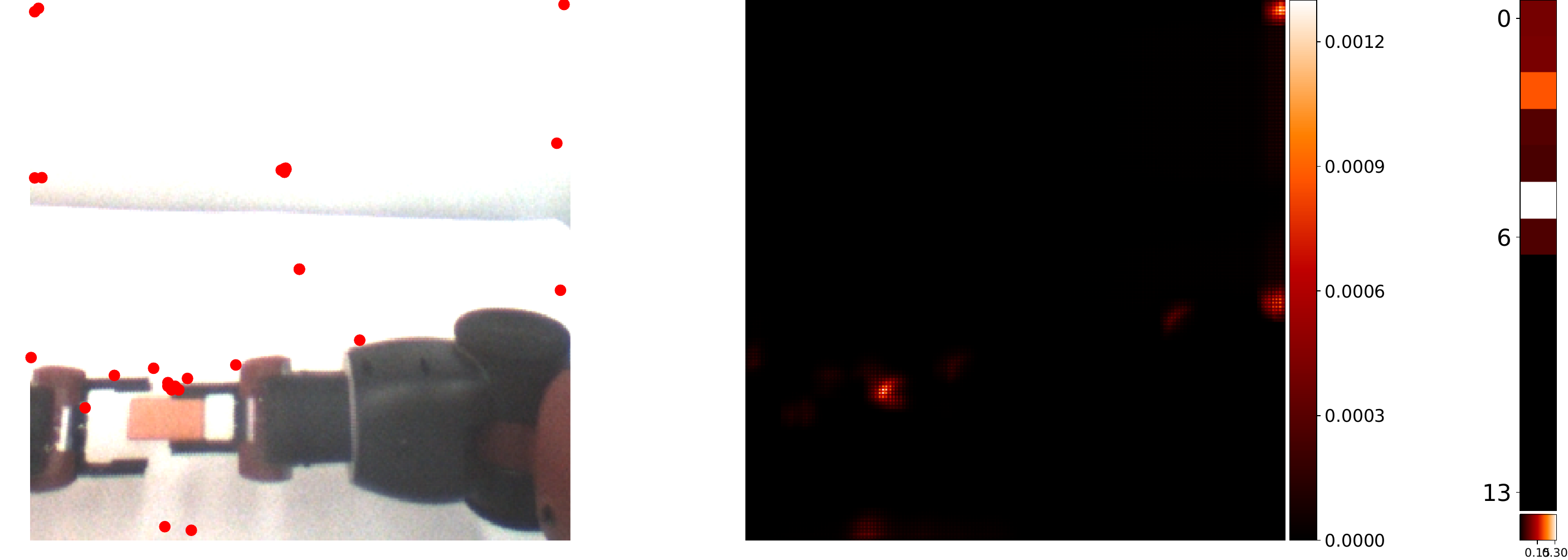}
    \caption{A heatmap in the Baxter reaching task}
    \label{fig:heatmap_baxter_reacher}
    \end{subfigure}
    \caption{Visualization results of qualitative analysis. Visual input with feature points is on the left of each figure with its corresponding heatmap, and configuration input is on the right with its corresponding heatmap. Brighter color represents high relevance score which means that corresponding input feature contributes more to the decisions of the robot policy.}
    \label{fig:heatmap_mjc_gps}
\end{figure*}

In this paper, we solve this problem by taking the absolute to the negative output before redistributing the relevance score. To this end, the signs of weights which are connected to the negative output are flipped. It only changes the direction of the relevance and preserves the absolute amount of relevance. We can interpret the propagated relevance from the absolute output as the absolute amount of the contribution of the last hidden layer to the output layer. Thus, the relevance at the last layer is propagated from Equation \ref{eq:last_rel},
\begin{align}
	\label{eq:last_rel}
	R_i = \sum_j \Big( \mathbb{1}_{\{\sum_{i'} z_{{i'}j} > 0\}} \frac{z_{ij}^+}{\sum_{i'} z_{{i'}j}^+} - \mathbb{1}_{\{\sum_{i'} z_{{i'}j} < 0\}} \frac{z_{ij}^-}{\sum_{i'} z_{{i'}j}^-} \Big) R_j,
\end{align}
where $z_{ij}^+$ and $z_{ij}^-$ are defined as $x_i w_{ij}^+$ and $x_i w_{ij}^-$ respectively. Following the definition from DTD, $w_{ij}^+$ is defined to be $\max(w_{ij}, 0)$ and $w_{ij}^-$ is defined to be $\min(w_{ij}, 0)$. By adding the second term, the relevance for the negative torque output is measured.

The existence of a negative value in the inputs, such as a negative joint position or negative joint velocity, is also problematic when applying the $z^+$-rule because it is derived from the constraints of positive inputs. If a negative value exists in the inputs, the root point is not guaranteed to be found. To address this problem, the sign of the negative input and the sign of their connected weights are flipped. This guarantees that the root point can be found, while also preserving the function value. Thus, the relevance is propagated to the layer which includes the negative input from Equation \ref{eq:input_rel},
\begin{align}
	\label{eq:input_rel}
    R_i = \sum_j \frac{\mathbb{1}_{\{x_i > 0\}} z_{ij}^+ + \mathbb{1}_{\{x_i < 0\}} z_{ij}^-}{\sum_{i'}(\mathbb{1}_{\{x_{i'} > 0\}} z_{i'j}^+ + \mathbb{1}_{\{x_{i'} < 0\}} z_{i'j}^-)}.
\end{align}

The second term propagates the relevance $R_j$ to negative input features by flipping the signs of the negative input and their connected weights. Finally, we achieve the relevance of each input feature by adding the relevance values of all joint torques, which are weighted by the importance factor $\alpha$.


\section{Experimental Results}
\subsection{Experimental Settings}
\label{sec:experimental_settings}
In our experiments, we analyze the decisions made from deep visuomotor policy models trained on simulated systems as well as real-world systems. We measure relevance of each input feature to the total end-effector movement. MuJoCo arms are used for the simulated systems \cite{mujoco}, and the Baxter research robot is used for the real-world systems. In the simulations, three different tasks are assigned to examine how the relevance of each input changes depending on the task. The first of these is a reaching task with a 2-DoF MuJoCo arm. The goal is to reach the target object by observing a raw image, where four different target points are given. For this task, the positions and velocities of both joints and the end-effector are given as a configuration input to the policy network. The second task is a peg insertion task with a 7-DoF arm. Initially, the arm is located at four different positions with the peg held in its gripper. The goal is to insert the peg into a hole when the robot observes a raw image of its arm and the hole. As the configuration input, joint positions and joint velocities are given. The final task is a door opening task. The goal of this task for the robot is to open a door with a hook attached to its end-effector. Similar to the previous task, the simulated MuJoCo agent has a 7-DoF arm. It is also trained to accomplish the task in four different conditions, where the initial location of the agent is different.

In real-world systems, the reaching task is trained with the Baxter research robot. Similar to the reaching task in the simulation setting, the goal is to reach the target, but in this case, the target location to reach is the position of the end-effector of the left arm. The agent is considered to accomplish the task if the block held by the right end-effector reaches to the left end-effector. No configuration information about the left arm, such as the joint position or end-effector position, is provided. A raw observed image and joint configurations which consist of joint positions and joint velocities of the right arm at each time step are provided as the input of the policy model.

\subsection{Qualitative Analysis of Deep Policy Networks}
\label{sec:qualitative}
We qualitatively analyze the policy models by visualizing a heatmap using DTD. For the multiple different experimental settings, we aim to answer the following questions: Does the inside of the deep policy network work as expected? Does it really refer to the image feature that is related to the task? Does observing the relevant image feature affect the performance and/or the task completion?

The visualization results from DTD are presented in Fig. \ref{fig:heatmap_mjc_gps}. Feature points extracted from the vision layer are plotted on the observed raw image in the image on the left in Fig. \ref{fig:heatmap_mjc_gps}. The heatmaps shown on the right side of each robot image present the relevance score of the observed image per pixel and the relevance score of the joint configuration.

In the case of the reaching task shown in Fig. \ref{fig:heatmap_mjc_reacher}, some feature points which are extracted from the vision layer \cite{gps_visuomotor} are located at unexpected positions. For example, feature points located on the top middle and bottom right appear to be irrelevant with regard to accomplishing the task for humans, although most of them are on the target object. In fact, the model is affected by these two task-irrelevant points, as as shown in the heatmap where the relevance score is high at these points. In the case of the peg insertion task shown in Fig. \ref{fig:heatmap_mjc_peg}, more unexpected feature points exist than in the previous task. Although the relevance at the feature points which are on the robot arm is higher than almost all other feature points, the relevance at the feature point located on top of the desk is also high, which was unexpected. A heatmap of the door opening task is presented in Fig. \ref{fig:heatmap_mjc_door}. In this task, there are even more task-irrelevant feature points. Even worse, the relevance at these points is higher than those at the others. However, surprisingly, the agent succeeds in accomplishing all tasks. In the case of the real-world robot system shown in Fig. \ref{fig:heatmap_baxter_reacher}, the policy network basically refers to the location of the block and the robot arm. However, it is also affected more by features at the top right and at the bottom of the image; these appear to be irrelevant with regard to how a human solves the task.

The experimental results from the qualitative analysis show that the robot policy models are affected more by task-irrelevant image features as the complexity of the task increases. However, this does not always result in poor task completion. This implies that referring to task-relevant image features is not a necessary condition for solving the task. We expand the interpretation from the results of qualitative analysis further in the following sections with the results of a quantitative analysis.
\subsection{Quantitative Analysis of Deep Policy Networks}
\label{sec:quantitative}
This section provides a quantitative analysis of which input feature attributes more to the manipulation of joints in robotic arms. For multiple different experimental settings, we aim to answer the following questions: To what extent does the deep policy model rely on two different inputs: the visual input and the configuration input? How does the relevance ratio change during the execution of the policy? Which role is each input feature responsible for?

\subsubsection{Static Analysis of Robotic Manipulations}
\label{sec:static}
\begin{figure}
    \centering
    \includegraphics[width=\linewidth]{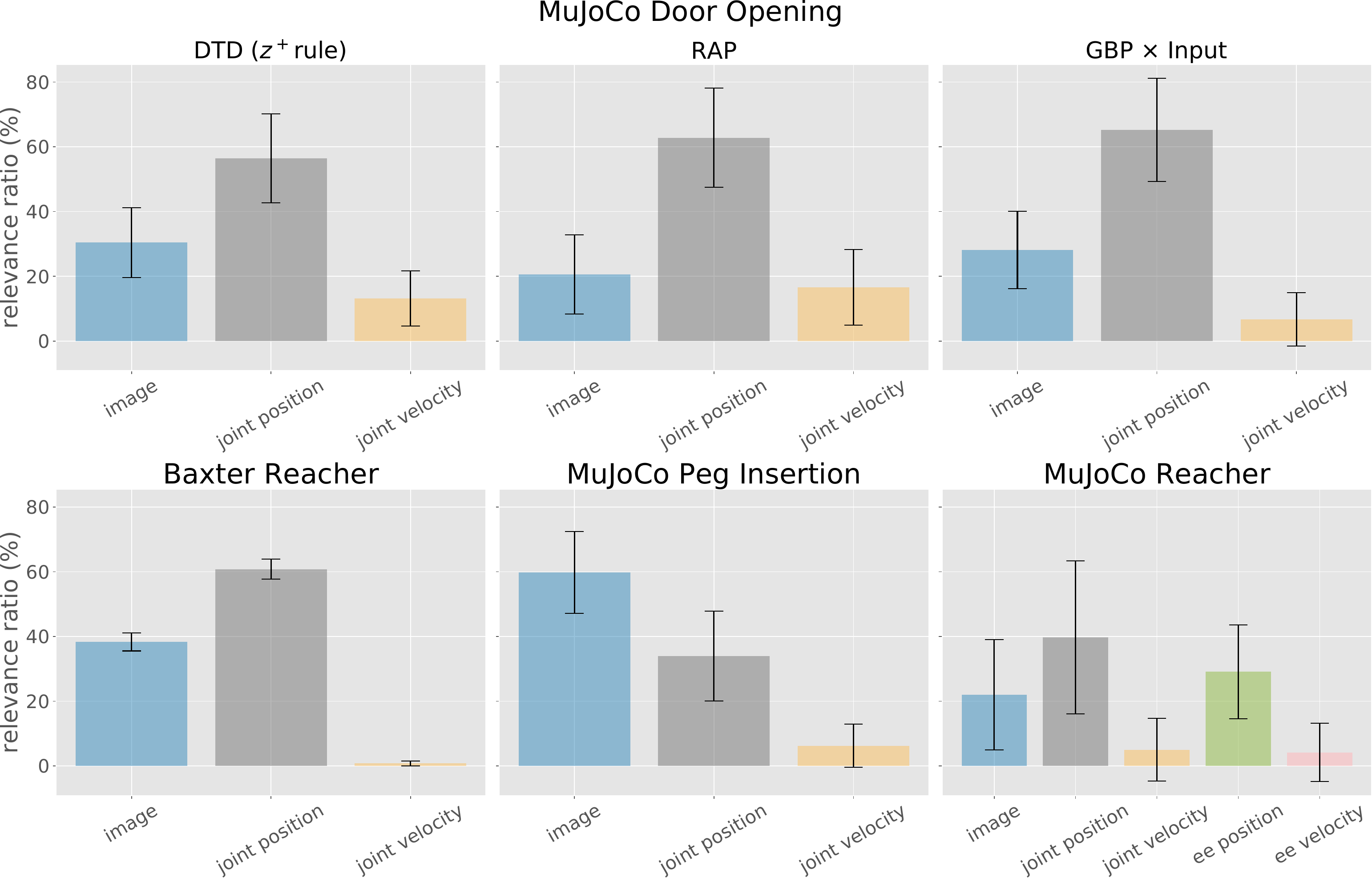}
    \caption{Bar chart of the relevance ratio of the MuJoCo door opening task (top), and bar chart of the relevance ratio of the Baxter reaching task, the MuJoCo peg insertion task, and the reaching task from DTD (bottom). Each bar represents the mean of the relevance ratio of each input feature with the standard deviation along the trajectory of the policy.}
    \label{fig:bar_mjc_gps}
\end{figure}

We analyze which input feature contributes more to the motor torque to understand the features which features we have to consider when we design the model. In order to verify the correctness of the relevance, we apply three methods: DTD, RAP, and GBP. Fig. \ref{fig:bar_mjc_gps} represents the average relevance ratio of each input feature along the trajectory of the policy. The relevance ratio from RAP and GBP in the MuJoCo reaching task, the peg insertion task, and the Baxter reaching task are not presented due to space constraints.

In Fig. \ref{fig:bar_mjc_gps}, the results show a consistent tendency of the average relevance ratio of each input feature for each task among all methods. The slight differences in the results come from a different propagating rule of each method. For example, the DTD $z^+$ rule masks negative weights $w_{ij}^-$ to zero and propagates the relevance, while GBP propagates the relevance and zeroes out negative gradients.

In the MuJoCo reaching task, the position information is considered to be the most important feature, while the image information is the next. Generally, three features, the image, joint position, and end-effector position, contribute to the decisions of the motor torque by a rate exceeding 90$\%$. In the case of the MuJoCo peg insertion task, the relevance ratios from each input attribution method differ slightly from each other. Nevertheless, both the image and position input features are considered to be important for solving the task.
Similarly, in the case of the door opening task and the Baxter reaching task, the joint position is said to be the most important feature by all methods. The second most important feature is the image feature. The policy model refers to the velocity feature the least.

In general, the position information, including the joint position and end-effector position, is considered to be the most valuable input, while the velocity information is considered to be the least valuable input in all tasks. In summary, the ratios of the relevance of the image input and the relevance of the configuration input are approximately 20:80, 60:40, 30:70, and 35:65 in the MuJoCo reaching task, the peg insertion task, the door opening task, and the Baxter reaching task, respectively.

\subsubsection{Dynamic Analysis of Robotic Manipulations}
\label{sec:dynamic}
\begin{figure}
    \centering
    \includegraphics[width=\linewidth]{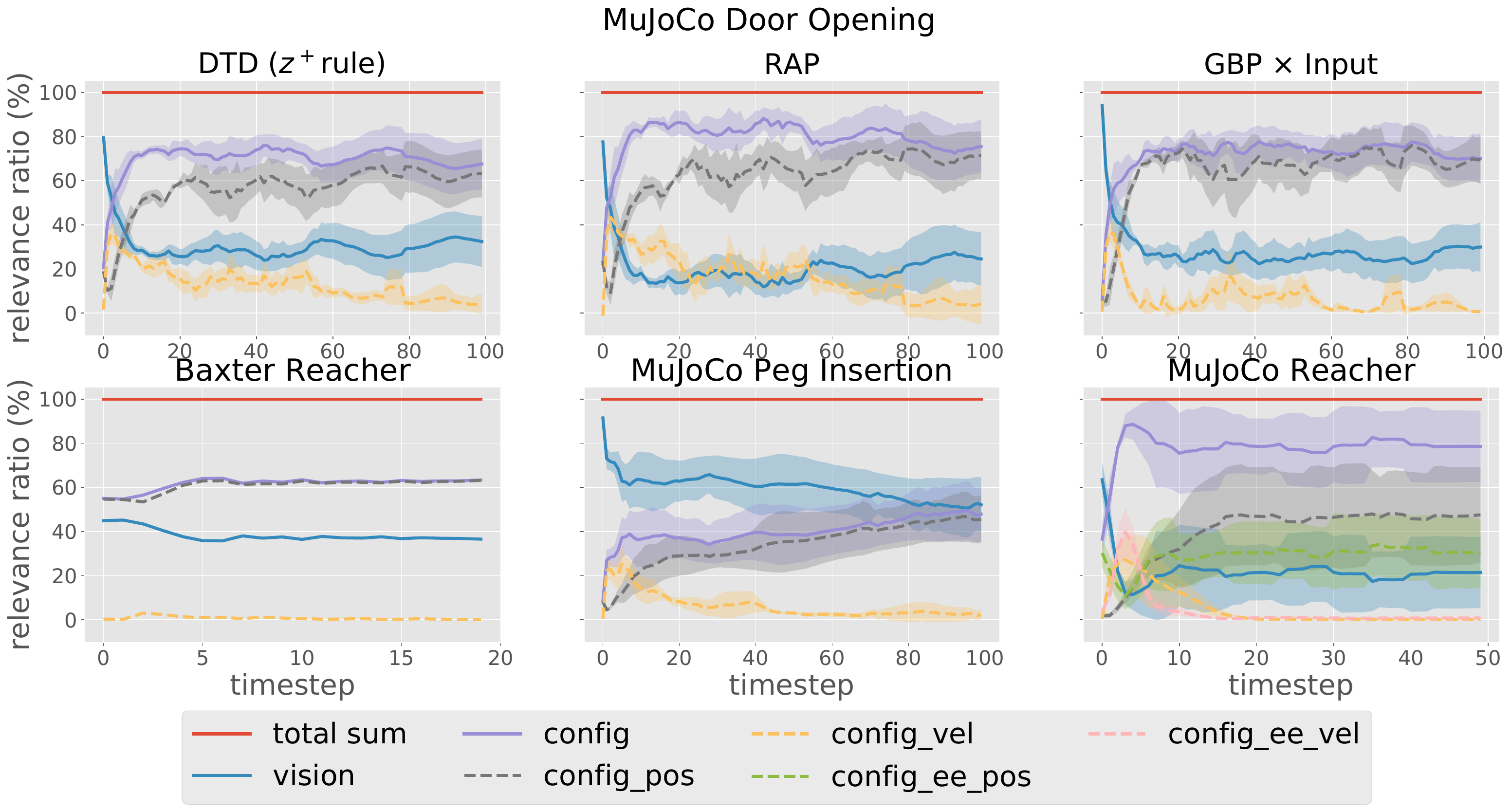}
    \caption{Relevance ratio change of the MuJoCo door opening task (top), and the Baxter reaching task, the MuJoCo peg insertion task, and the reaching task with DTD ($z^+$-rule) (bottom). Red, blue, and purple solid lines represent the ratio of the total relevance, image relevance, and configuration relevance, respectively. Black, yellow, green, and pink dotted lines represent the joint position, joint velocity, end-effector position, and end-effector velocity, respectively.}
    \label{fig:ratio_mjc_gps}
\end{figure}

This section analyzes how the relevance ratio changes as the policy executes. Fig. \ref{fig:ratio_mjc_gps} represents the relevance ratio change during the execution of the policy. The trends of the relevance ratio change results from the three different methods show consistency in each task. For example, when the relevance of the configuration input increases at time step 1 in the case of the peg insertion task from DTD, the relevance of the configuration input from RAP and GBP at that time step also increases.


Surprisingly, the tendencies of the relevance ratio changes are similar across all tasks, despite the fact that an individual policy model is trained for each task. The relevance of the image in the initial step is significantly high, and it declines considerably. On the other hand, position information in the initial step is not considered as an important feature. However, the relevance of the joint position (and the end-effector position if it exists) increases steeply. Similar to the image input, the relevance of the joint velocity (and end-effector velocity if it exists) falls considerably as the time step increases.

Such a situation is necessary to differentiate where the target is. As we describe in Section \ref{sec:experimental_settings}, in all tasks, the agent is trained to accomplish the given task under several different conditions. Different conditions could be the different initial position of the agent or a different position of the target object. Note that both have the same effect, as the base coordinate is defined on the torso of the robot. In this regard, the position input does not provide any clue as to where the target is at the initial time step. Therefore, at the beginning, the visuomotor policy model refers to the image information to determine where the target is.

\begin{figure}
    \centering
    \includegraphics[width=\linewidth]{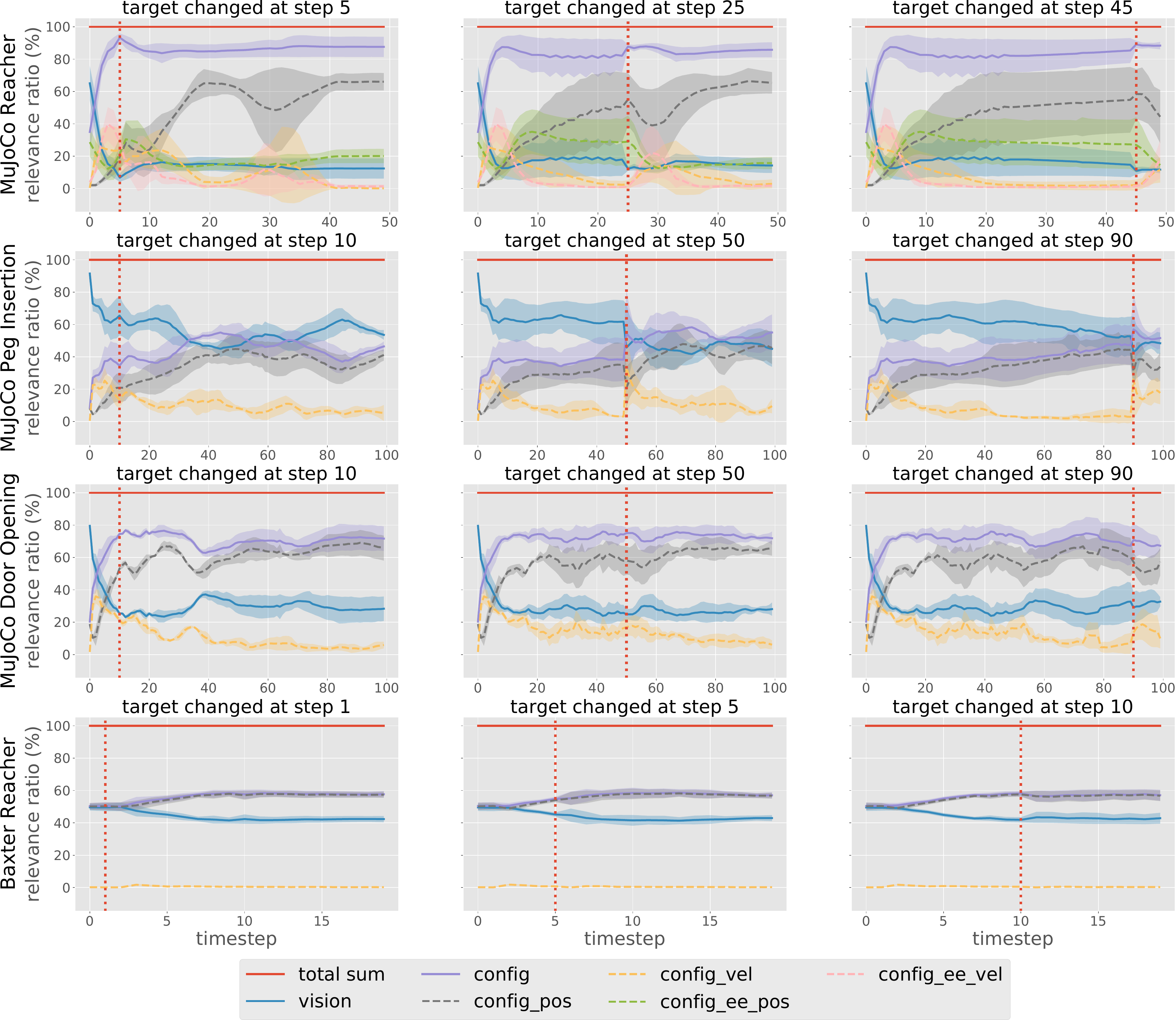}
    \caption{Relevance ratio change when the target is changed. Red, blue, and purple solid lines represent the ratio of the total relevance, image relevance, and configuration relevance, respectively. Black, yellow, green, and pink dotted lines represent the joint position, joint velocity, end-effector position, and end-effector velocity, respectively. The time step when the target is changed is represented by the vertical dotted line.}
    \label{fig:ratio_target_change}
\end{figure}

Especially for the GPS algorithm that we used for the training algorithm, the policy model is trained to follow the mean of a Gaussian trajectory, which is determined by solving the trajectory optimization problem \cite{gps_traj_opt}. In this procedure, a Gaussian trajectory is generated only from the joint configuration. Therefore, the variance in early time steps is high given that the robot arm has a similar joint configuration in these time steps for each different condition. However, the variance decreases as it approaches the target. For this reason, when the variance of the trajectory is high in the initial step, the trajectory is selected by considering the image input. However, as the variance decreases, the position information becomes more valuable as using it is sufficient to determine which trajectory current policy follows with the joint position. Therefore, the model starts to consider the position information more as the policy runs.
\subsubsection{Dynamic Analysis of Robot Behavior when the Target Changes}
\label{sec:dynamic_target_change}
This section analyzes how the inside of the model reacts when a perturbation occurs. To this end, the dynamic change of the relevance ratio is observed while the target is changed in the middle of the policy execution. The changed target position is randomly located in between the trained target positions. The relevance ratio results in this section are computed from the DTD $z^+$-rule, as this rule has a solid theoretical background.

In Fig. \ref{fig:ratio_target_change}, we show the relevance ratio change results when the target is changed. The red vertical dotted line represents the time step when the target is changed.
In the previous section, we find that the relevance ratio of the position feature increases constantly, whereas the relevance ratio of the velocity feature decreases. However, when the target is changed in the middle of the policy execution, the relevance of the position feature suddenly drops. It then bounces back after several time steps from the target change. On the other hand, the policy model refers more to the vision information and the velocity information immediately after the target change occurs. After several time steps from the target change, the relevance ratio of the velocity decreases again.

These results imply that position information plays an important role in local manipulation because the relevance ratio of the position feature increases as the robot arm moves closer to the target object. In contrast, the relevance ratio of the velocity feature peaks in the early stage of policy execution when the robot arm is far from the target object, but then drops as it becomes closer to the target object. These results imply that velocity information is in charge of global manipulation. Meanwhile, visual information provides a clue in the early stage about which trajectory it should follow.
\section{Conclusion}
\label{conclusion}
In this paper, we explain the decisions of the deep visuomotor policy models for robotic manipulation. In order to analyze the decision making reasons, three different input attribution methods, Deep Taylor Decomposition, Relative Attributing Propagation, and Guided Backpropagation, are utilized. To handle negative inputs and outputs properly, we propose a modified relevance propagation method. In addition, we expand input attribution methods which were originally proposed for image classifiers to explain robot policy networks by computing the contributions of each torque on the end-effector movement. With these methods, we explicitly measure the extent to which each input contributes to the decisions of the policy models and visualize a heatmap. From the results of a qualitative analysis, we identify that referring to task-relevant image features is not a necessary condition for task completion. In addition, the dynamic relevance ratio change results imply that position information is responsible for local manipulation, while velocity information is responsible for global manipulation. Meanwhile, the policy model chooses the trajectory to follow by referring to visual information in the initial stage of policy execution. To the best of our knowledge, this is the first report to identify the dynamic changes of the input attributions of sensor inputs in deep policy networks for robotic manipulation.

\addtolength{\textheight}{-12cm}   






\section*{ACKNOWLEDGMENT}
This work was supported by Institute of Information \& communications Technology Planning \& Evaluation (IITP) grant funded by the Ministry of Science and ICT (MSIT) (No. 2017-0-01779, XAI and No. 2019-0-00075, Artificial Intelligence Graduate School Program (KAIST)) and Industrial Strategic Technology Development Program funded by the Ministry of Trade, Industry \& Energy (MOTIE, Korea) (No. 10077533, Development of robotic manipulation algorithm for grasping/assembling with the machine learning using visual and tactile sensing information).

\bibliographystyle{IEEEtran}
\bibliography{root}

\end{document}